\documentclass[letterpaper]{article} 
\usepackage[submission]{aaai24}  
\usepackage{times}  
\usepackage{helvet}  
\usepackage{courier}  
\usepackage[hyphens]{url}  
\usepackage{graphicx} 
\usepackage{natbib}  
\usepackage{caption} 
\usepackage{algorithm}

\usepackage{newfloat}
\usepackage{listings}
\usepackage{cite}
\usepackage{amsmath,amssymb,amsfonts}
\usepackage{graphicx}
\usepackage{textcomp}
\usepackage{xcolor}
\usepackage{algorithm}
\usepackage[noend]{algpseudocode}
\algnewcommand{\Or}{\textbf{ or }}
\algnewcommand{\And}{\textbf{ and }}
\algnewcommand{\Not}{\textbf{not }}
\usepackage{multirow}
\usepackage{multicol}
\usepackage{amsmath}
\usepackage{graphics}
\usepackage{graphicx}
\usepackage{todonotes}
\usepackage{subfig}
\DeclareCaptionStyle{ruled}{labelfont=normalfont,labelsep=colon,strut=off} 
\floatstyle{ruled}
\newfloat{listing}{tb}{lst}{}
\floatname{listing}{Listing}
\title{Uncertainty in Additive Feature Attribution methods}

\author{
    Abhishek Madaan \equalcontrib \textsuperscript{\rm 1},
    Tanya Chowdhury \equalcontrib \textsuperscript{\rm 2},
    Neha Rana \textsuperscript{\rm 1},
    James Allan \textsuperscript{\rm 2},
    Tanmoy Chakraborty \textsuperscript{\rm 3},
}
\affiliations{
    \textsuperscript{\rm 1}IIIT-Delhi, New Delhi, India\\
    \textsuperscript{\rm 2}University of Massachusetts Amherst, USA\\
    \textsuperscript{\rm 3}IIT Delhi, New Delhi, India\\

}

\usepackage{bibentry}

\begin{document}
\maketitle
\begin{abstract}
In this work, we explore various topics that fall under the umbrella of Uncertainty in post-hoc Explainable AI (XAI) methods. We in particular focus on the class of additive feature attribution explanation methods. We first describe our specifications of uncertainty and compare various statistical and recent methods to quantify the same. Next, for a particular instance, we study the relationship between a feature's attribution and its uncertainty and observe little correlation. As a result, we propose a modification in the distribution from which perturbations are sampled in LIME-based algorithms such that the important features have minimal uncertainty without an increase in computational cost. Next, while studying how the uncertainty in explanations varies across the feature space of a classifier, we observe that a fraction of instances show near-zero uncertainty. We coin the term ``stable instances'' for such instances and diagnose factors that make an instance stable.  
Next, we study how an XAI algorithm's uncertainty varies with the size and complexity of the underlying model. We observe that the more complex the model, the more inherent uncertainty is exhibited by it.  As a result, we propose a measure to quantify the relative complexity of a blackbox classifier. This could be incorporated, for example, in LIME-based algorithms' sampling densities, to help different explanation algorithms achieve tighter confidence levels. Together, the above measures would have a strong impact on making XAI models relatively trustworthy for the end-user as well as aiding scientific discovery. 
\end{abstract}

%
\begin{figure*}[!t]
    \centering
    \subfloat{\includegraphics[width=0.45\linewidth]{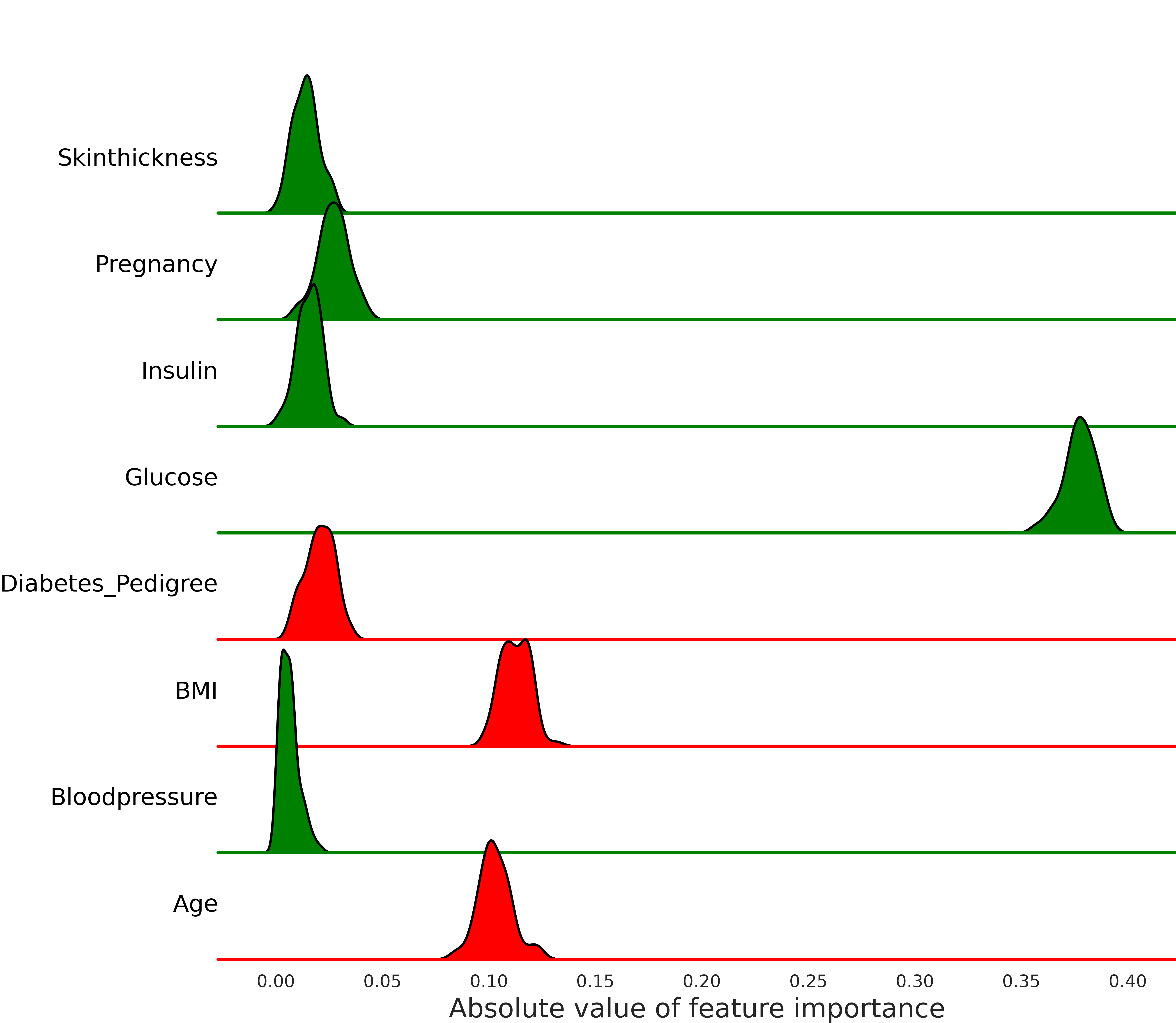} }%
    \qquad
    \subfloat{\includegraphics[width=0.45\linewidth]{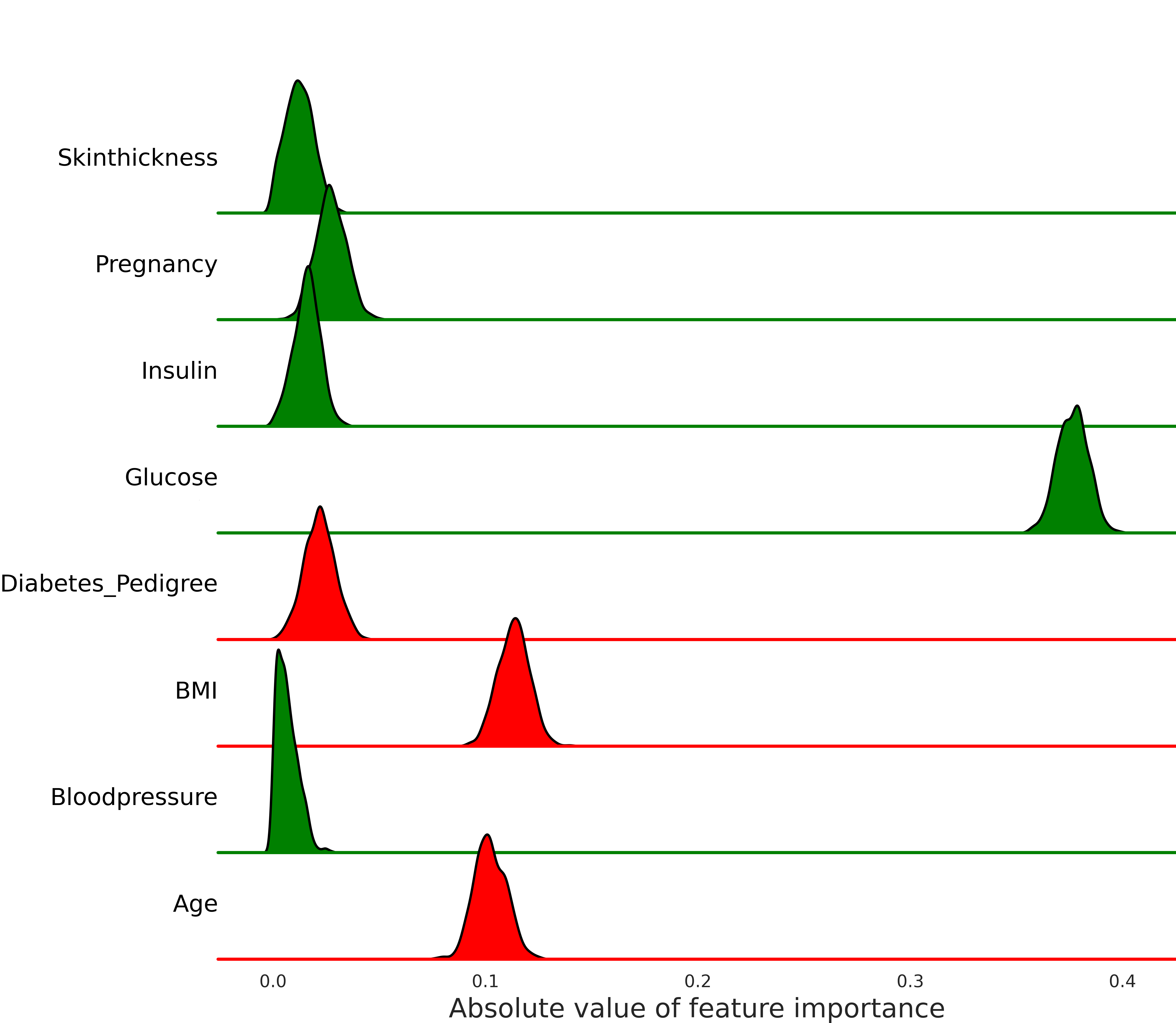} }%
    \caption{Feature-wise distribution of variation in importance scores obtained by executing the LIME explanation algorithm (left) 100, and (right) 1000 times, respectively to explain a binary classifier on a chosen instance of the UCI Diabetes dataset.}%
    \label{fig:unc100}%
\end{figure*}

\section{Introduction}
The last couple of years have witnessed the advent of a large number of feature attribution methods \cite{ribeiro2016should,lundberg2017unified,shrikumar2017learning}. While some of them are motivated by generating explanations of deep neural models for end users (model-based explanations), others are built in order to gain an understanding of unknown properties and mechanisms of the underlying data-generating process (system-based explanations).  Many feature attribution algorithms have been independently evaluated for stability/robustness/uncertainty, and unfortunately, most of them fare poorly \cite{schulz2021uncertainty}. There is a pressing need to quantify and reduce uncertainty of these explanation algorithms. For end users, upper bounding the uncertainty estimate is necessary for garnering trust on the use of Explainable-AI models. It is especially important to do so before they are used in critical decision-making scenarios such as deciding jail terms, making medical diagnoses,  etc. System-based explanation methods consider the explanation algorithm to be genuine insights, so even small errors in the algorithm's results can lead to false discoveries and incorrect scientific understanding. Uncertainty quantification is critical to address that issue. 

Only a handful of works have attempted to address problems related to uncertainty in feature attribution methods. Certain XAI algorithms report the uncertainty associated with a feature attribution. For example: BayesLIME ~\cite{slack2021reliable} computes instance-wise uncertainty at various confidence levels;  CXPlain \cite{schwab2019cxplain} reports global uncertainty for the explanation algorithm. \citet{ghorbani2019interpretation} point out how explanation algorithms are stochastic processes, highly susceptible to hyper-parameter changes and adversarial feature perturbations. \citet{zhang2019why} focus on the LIME algorithm and diagnose the three different root causes of uncertainty in the same while \citet{schulz2021uncertainty} attempt to assign an uncertainty estimate to LIME feature attributions using bootstrapping and ordinal consensus methods.
\citet{krishna2022disagreement} study the \textit{disagreement problem} among feature attributions and propose interesting metrics such as Feature agreement and Rank agreement that can be used to quantify uncertainty among feature attribution methods. 
However the problem space of feature attribution algorithms and uncertainties is huge and the research done so far is in its  early stages.
Owing to the limited work in this domain and the sheer importance of the research problem, we conduct a behavioral exploration on various research questions under the umbrella of Uncertainty in XAI. Based on our findings, we also suggest interventions which would make existing algorithms considerably more robust through the lens of uncertainty.

In this work, we analyse the uncertainty presented by post-hoc additive feature attribution based XAI methods for generating local explanations. We first describe our definition and set of assumptions regarding XAI uncertainty and enumerate statistical and recent measures to quantify the same (Confidence Interval w/o Bootstrapping, Standard Deviation, Kendall's Coefficient of Concordance, Fleiss's Kappa, Feature Agreement, and Rank Agreement). We tabulate and compare the uncertainty reported by each metric on four popular feature attribution methods (LIME, KernelSHAP, BayesLIME, and CXPlain) over two datasets. Because of its balance between stability and sensitivity, we choose the Kendall's coefficient of concordance to quantify uncertainty in our experiments.
We next study the relationship between a feature's attribution and its uncertainty, and observe the absence of any strong correlation. We propose a modification in the multivariate Gaussian distribution from which perturbations are sampled in the LIME algorithm such that the important features have $11.42\%$ reduced uncertainty as compared to the original algorithm without an increase in computational cost. While studying how uncertainty varies across the feature space of a dataset, we witness that a fraction of instances show near-zero uncertainty; we coin the term ``stable instances'' for such instances. We diagnose which factors lead to an instance being stable, and propose an algorithm that determines if an instance is stable with a precision of $89.2\%$ and recall of $95.1\%$.
Next, we study how uncertainties of XAI algorithms  vary with the size and complexity of the underlying models. We observe that the more complex the model, the more the inherent uncertainty exhibited by it.  As a result, we propose an important measure to quantify the complexity of a blackbox model. This could be incorporated in numerous applications, for example, in LIME-based algorithms to decide sampling densities, to aid different models achieve given confidence levels.

{\bf Our major contributions are as follows:}\\
$\bullet$ We benchmark the performance of 4 different XAI algorithms on 5 uncertainty quantification metrics for a small-scale diabetes and a larger-scale synthetic dataset.
$\bullet$ We propose a modification to the multivariate sampling function of LIME to achieve $11.42\%$ and $9.06\%$ reduction in top attributes over the PIMA Diabetes and synthetic datasets, respectively.
$\bullet$ We coin the term ``stable instances'' for instances whose uncertainty remains low across executions and propose an algorithm to mine ``stable instances''.
$\bullet$ We find a strong correlation between the ``complexity'' of the underlying model and its average uncertainty and propose a cardinality-based metric to approximate the complexity of a model for explanation computation.

To the best of our knowledge, we are the first to study the aforementioned research questions for XAI. In the spirit of reproducability, we share all the scripts and datasets required as supplementary material.

\begin{table*}[!t]
\centering
\caption{Measures (Confidence Interval (CI), Confidence Interval Bootstrapping (CI-bootstrap), Standard Deviation (Std. dev.), Kendall's Coefficient of Concordance ($W$), Fleiss' Kappa ($\kappa$), Top-K Feature Agreement (TopK-FA) and Top-K Rank Agreement (TopK-RA) with k=8) to quantify uncertainty in explanations  generated using popular local linear explanation methods (LIME, KernelShap, BayesLIME and CXPlain) using the PIMA Indian diabetes and our synthetic dataset. }

\begin{tabular}{l|l|l|l|l|l|l|l}
\hline
  Metric & CI & CI-bootstrap & Std. dev. & W & $\kappa$ & TopK-FA & TopK-RA \\ \hline \hline
  \multicolumn{8}{c}{Diabetes}\\\hline 
    Random  & 2.546& 2.156& 4.120 & 0.987&0.999 & 0 &0.875 \\  \hline
  LIME  & 0.005  &  0.005   &0.014 & 0.160 & 0.661 &0 &0.5\\ \hline
  KernelShap & 0.0043  & 0.0042 & 0.0108 & 0.058& 0.326 &0 & 0.375\\ \hline
  BayesLIME  & 0.012 &0.0121  &0.031   & 0.862  &  0.473 & 0 & 0.75\\ \hline
  CXPlain & 0.0402 & 0.0327 & 0.0482 & 0.546 & 0.433 & 0 & 0\\
  \hline
  \hline
  \multicolumn{8}{c}{Synthetic} \\\hline 
  Random  & 2.783& 2.929& 3.997 &1 &0.990 & 0.892 & 0.979\\  \hline
  LIME  &  $1.64\times10^{-3}$ & $1.62\times10^{-3}$   & $4.11\times10^{-3}$ & 0.569 &0.953 & 0.465 & 0.894 \\\hline
  KernelShap & $1.11\times10^{44}$   & $8.38\times10^{43}$ & $2.99\times10^{45}$ & 0.991  & 0.9995 & 0.35 & 0.875  \\ \hline
  BayesLIME  &0.035  &0.034  &0.088  &0.971  &0.906 & 0.234 & 0.76   \\ \hline
  CXPlain & 0.0123 &  0.0009 & 0.0012 &  0.0685& 0.939 & 0 & 0.155\\
  \hline
\end{tabular}
\label{table : table1}
\end{table*}

\section{Dataset Descriptions}
We use two datasets for our experiments: the \textit{UCI Pima Indians Diabetes dataset}\footnote{\url{https://raw.githubusercontent.com/jbrownlee/Datasets/master/pima-indians-diabetes.csv}} and a larger scale synthetic dataset. 

Given certain features from patient medical records, the \textit{Pima Indians Diabetes} dataset is a binary classification dataset that predicts the onset of diabetes within 5 years. It has 768 observations, containing 8 input features (Glucose, Blood Pressure, Insulin, BMI, Age, etc.) and a binary output where 1 indicates eventual disease onset.  The train-test split for this dataset is $80:20$. This dataset choice was motivated by the large use case of XAI tools in medical diagnostic settings and the urgent need to curb uncertainty in such scenarios.

To better understand underlying phenomena of XAI methods, we validate our experiments on a synthetic dataset. We build a dataset with $100$ features and $200,000$ instances for our experiments. To build our dataset, we focus on generating two independent parts: instances $x$ and their ground-truth labels $y$.  We sample our instances $x$ from a mixture of multivariate Gaussians as described by \citet{liu2021synthetic} starting with arbitrary values of $\mu_D$ and $\Sigma_{D\times D}$ (here $D=100$). For generating ground-truth labels, we implement a set of linear functions, piece-wise constant functions, non linear additive functions (e.g., absolute, sine, cosine, exponent, etc.) and piece-wise linear functions.  We then compute intermediate labels $y_{raw} = \sum_{n=1}^{100} \Psi_n (x_n)$, where $\Psi_n$ refers to one of the functions described above. The labels obtained in this manner are normalized to have zero mean and unit variance. To build the binary classification dataset, instances whose $y_{raw}$ values lie above the mean are assigned label $1$, while those below are assigned label $0$. We set the train-test split for this dataset at $75:25$ \cite{liu2021synthetic}. This synthetic dataset allows us to perform targeted experiments by providing control of each attribute and enables the simulation of various scenarios like low and high density regions, low and high correlation between features, as well as incorporating anomalous behaviour within a dataset.

\section{Quantifying Uncertainty in XAI} \label{sec:one}
While there can be multiple underlying causes that lead to uncertainty in feature attributions, like classifier uncertainty,  in this work we focus on uncertainty produced solely due to multiple runs of the same explanation algorithm. A large number of explanation algorithms are perturbation-based and as a result stochastic processes, whose uncertainty we wish to quantify. Next, we briefly enumerate methods that can be used to measure the uncertainty of feature attributions. We include statistical uncertainty quantification methods as well as variants proposed in recent literature specific to XAI. 

\noindent $\bullet$ \textbf{Confidence Interval}: Used by \citet{schwab2019cxplain} for XAI uncertainty quantification with the width of the confidence interval being the uncertainty measure.\\
$\bullet$ \textbf{Confidence Interval + Bootstrapping}: The attribution algorithm is run $M$ times before a median feature attribution is chosen from the ensemble on which \citet{schwab2019cxplain} compute uncertainty as above.\\
$\bullet$ \textbf{Standard Deviation}: The attribution algorithm is run $M$ times and the standard deviation (stdev) of each feature's attribution is calculated and averaged to compute an aggregated uncertainty score. \\
$\bullet$ \textbf{Kendall's Coefficient of Concordance  ($w$)}: Used by \citet{schulz2021uncertainty} in XAI,  to find correlation between orderings of features. This is a bounded metric and returns values in $[0,1]$. Since this is a concordance metric, $W = 1 -w$ is used to quantify uncertainty. \\
$\bullet$ \textbf{Fleiss' Kappa ($\kappa$)}: Introduced by \citet{fleiss1973equivalence} and also used by \citet{schulz2021uncertainty} in XAI, Kappa is also a rank correlation metric which returns a bounded value in $[0,1]$.\\
$\bullet$ \textbf{Feature Agreement}: Introduced by \citet{krishna2022disagreement}, this metric measures the relative agreement two attributions face on the top $k$ features. This score is pairwise aggregated over all attributions over which uncertainty is being evaluated. Since this is a similarity score, we report the score subtracted from 1 as uncertainty. \\
$\bullet$ \textbf{Rank Agreement}: Also introduced by \citet{krishna2022disagreement}, this metric computes the fraction of features that have the same position in respective rank orders in two feature attributions. This is a stricter metric than feature agreement.

\begin{table*}[!ht]
    \centering
    \caption{Comparing the uncertainty of the original LIME and KernelSHAP (KS) algorithms and the Multivariate Gaussian Sampled LIME (MVG-LIME) and MVG-KS modification we propose, over Top-1 feature, Top-2 feature and Top-5 feature and for the entire instance with k=2. While the first three uncertainties are computed using standard deviation, the instance uncertainty is computed using Kendall's coefficient of concordance. Results are averaged over $100$ instances from the Diabetes and synthetic dataset. }
    \begin{tabular}{c||c|c|c|c||c|c|c|c}
       \hline
      Metric  & LIME & MVG-LIME & KS & MVG-KS & LIME & MVG-LIME & KS & MVG-KS\\ \hline\hline  
      Dataset &  \multicolumn{4}{|c||}{Diabetes} & \multicolumn{4}{c}{Synthetic} \\ \hline
      Top-1 Unc (std)  & 0.011& 0.0092 & 0.012& 0.0095& 0.023& 0.0181& 0.036& 0.0067\\  
      Top-2 Unc (std) & 0.0117& 0.013 & 0.014& 0.0092& 0.026& 0.02& 0.039& 0.026\\  
      Top-5 Unc (std) & 0.014 & 0.0136 & 0.021& 0.0198& 0.032& 0.023& 0.043& 0.031\\  
      All Unc & 0.176 & 0.087 & 0.035& 0.0232& 0.065& 0.045 & 0.072 & 0.059\\ \hline
    \end{tabular}
    \label{tab:impunc}
\end{table*}

We perform experiments to quantify the uncertainty of four different explanation algorithms, namely LIME \cite{ribeiro2016should}, KernelSHAP \cite{lundberg2017unified}, BayesLIME \cite{slack2021reliable} and CXPlain \cite{schwab2019cxplain}, on the synthetic and the PIMA Indian diabetes dataset using the seven metrics to compute uncertainty described above. The experimental settings for this comparision along with details of each metric can be found in the Appendix. We report the comparision results in Table \ref{table : table1}. We first observe that all of the studied explanation algorithms show significant uncertainty, making the amortization of such uncertainty an important task. We also find that, among the discussed metrics, the Kendall's coefficient of concordance is bounded, assumes a range of values in the interval, as well as shows strong correlations to the other metrics. Based on that, we use the Kendall's Coefficient ($W$) to measure uncertainty in our remaining experiments.

\section{Feature Uncertainty vs Feature Importance} \label{sec:two}

 In this section, we focus on the uncertainty shown by individual features in a feature attribution. We witness significant uncertainty in each feature's attribution score (its ``importance'')  when running any of the above mentioned explanation algorithms multiple times with the same set of hyper-parameters (see Figure~\ref{fig:unc100}). It is of interest to us to study the correlation between a feature's importance and the uncertainty of its attribution. A negative correlation between them is desirable as most applications focus on the top K most important features ~\cite{lipton2018mythos}, for which we would like to minimize uncertainty.

As a result, we bootstrap and compute the uncertainty for each feature individually over  100 runs (Figure \ref{fig:unc100}). We then study the correlation between a feature's attribution ($w$) and its uncertainty ($u$) for a particular instance (Figure \ref{fig:diab_featimp_featunc} in Appendix). To do so, we compute the Pearson's correlation coefficient between $w_i$ and $u_i$:
$r = \frac{ \sum_i(w_i-\overline{w})(u_i-\overline{u}) }{ \sum (w_i - \overline{w})^2 (u_i - \overline{u})^2 }$.  We sample 100 instances from both datasets and compute the average Pearson correlation to demarcate the relationship between the feature's attribution and uncertainty in each feature. We find the mean correlation to be $0.128$ and $0.0104$ with a standard deviation of $0.015$ and $0.0006$ for the PIMA Diabetes and Synthetic datasets respectively. The Pearson's coeeficient has a range of $[-1,1]$, and our numbers show very weak positive correlation, which is not ideal. As a result, we propose a generic modification to perturbation-based explanation algorithms in order to reduce uncertainty of the important features.

We suggest a framework that allows important features to have stronger uncertainty bounds as compared to the less significant ones at the same sampling density. At present, when generating tabular perturbations, the LIME algorithm\footnote{\url{https://github.com/marcotcr/lime/blob/master/lime/lime_tabular.py}} accepts as input hyper-parameter \emph{num-samples} and samples a univariate isotropic Gaussian distribution to choose \emph{num-samples} instances in the neighborhood of the instance. This implied that perturbations are chosen around the instance $x$ being explained following the distribution 
$f_{\mathcal{N}}(x) = \frac{1}{\sigma \sqrt{2\pi}} e^{- \frac{(x-\mu)^2}{2\sigma^2}}$.
The algorithm sets $\mu$ to $0$ and $\sigma$ to $1$ by default. 

We know from \citet{laugel2018comparison} that an accurate approximation for the LIME explanation of an instance is the LIME explanation computed at its nearest Decision Boundary Point (DBP). A DBP for an instance is the instance on the classifier decision boundary with nearest Euclidean distance to the instance being explained. We first compute the tangent to the instance at its nearest DBP and obtain feature attributions $w_{DBP}$ (Appendix : Algorithm \ref{alg:growingSpheres}). We also measure the distance $l$ between the instance being explained and its DBP. We then use the weights of $w_{DBP}$ and the distance to the nearest DBP $l$ to sample perturbations in LIME in an updated covariance  matrix such that, $\mu_{new} = 0_D$, and $\Sigma_{new} = \frac{k\times w_{DBP}}{l}$, where $k$ is a constant, and $\Sigma_{new}$'s non-diagonal elements are zero. 

 We hypothesize that sampling from this multivariate Gaussian distribution with mean $\mu_{new}$ and covariance $\Sigma_{new}$ would lead to reduced uncertainty as compared to the original algorithm. We name this modified algorithm -- {\bf Multivariate Gaussian Sampled LIME} (MVG-LIME).

We experiment by generating LIME based explanations for 100 instances using both the classic method and our modification on both the datasets. We set the value of constant $k$ in MVG-LIME to $2$ on the diabetes dataset and $18$ on the synthetic dataset based on hyper-parameter tuning. All other hyper-parameters are fixed to the same values (default LIME parameters) across the experiments. The results are shown in Table \ref{tab:impunc}. We witness a $27.27\%$ reduction in the uncertainty of the top feature, $ 14.53\%$ in the top two features, and $7.386\%$ on the entire instance in the modified approach as compared to the base algorithm. 

This new sampling distribution enables features with higher attributions in the DBP to be sampled more densely as they are assigned a wider range. In a similar manner, features with low attributions are assigned a lower range for sampling and would not be as thoroughly sampled. Employing this modification would decrease the average uncertainty of important attributes in perturbation-based algorithms at the same computational cost.

\begin{figure*}[htp]
    \centering
    \subfloat{{\includegraphics[width=0.45\linewidth]{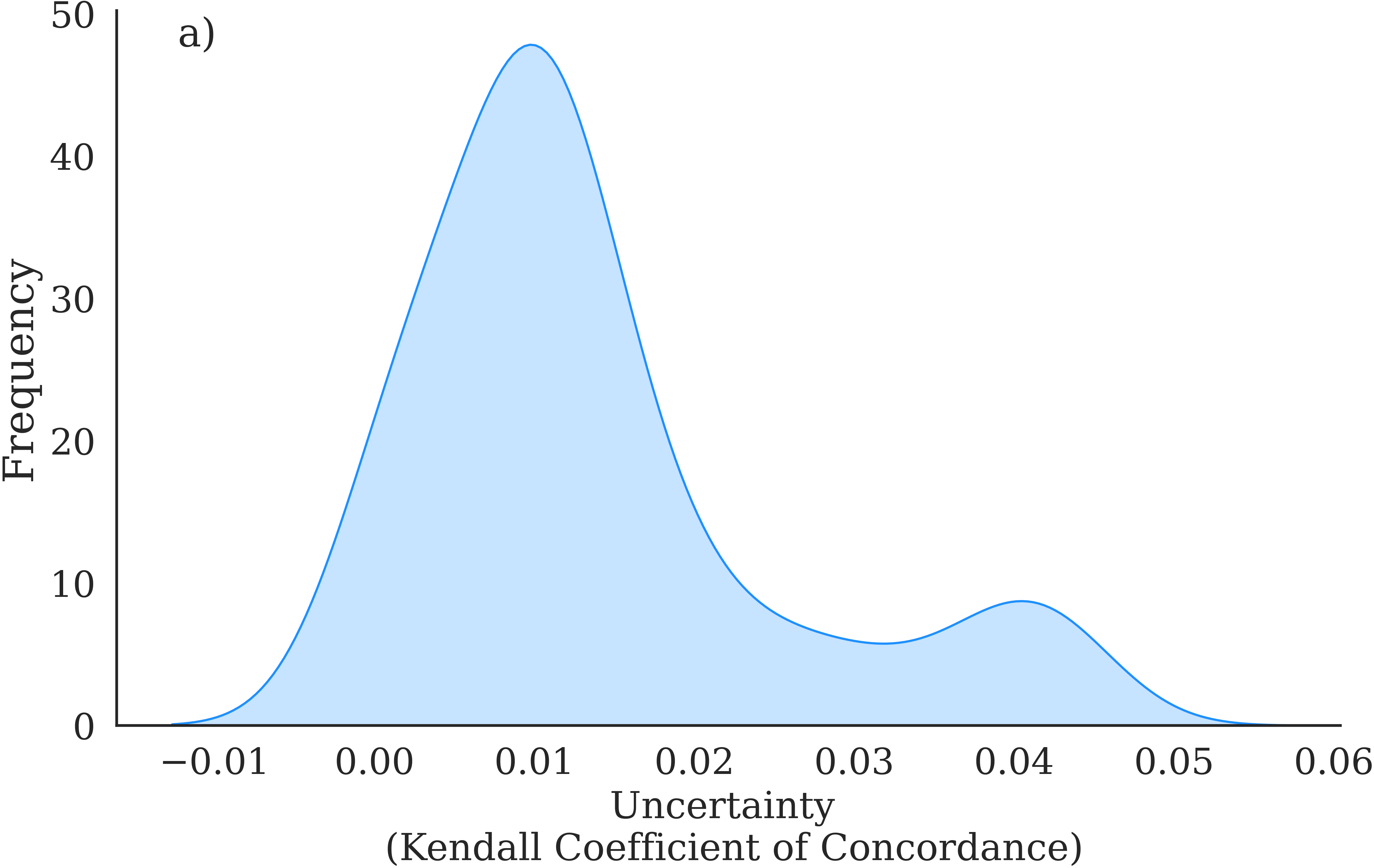} }}%
    \qquad
    \subfloat{{\includegraphics[width=0.45\linewidth]{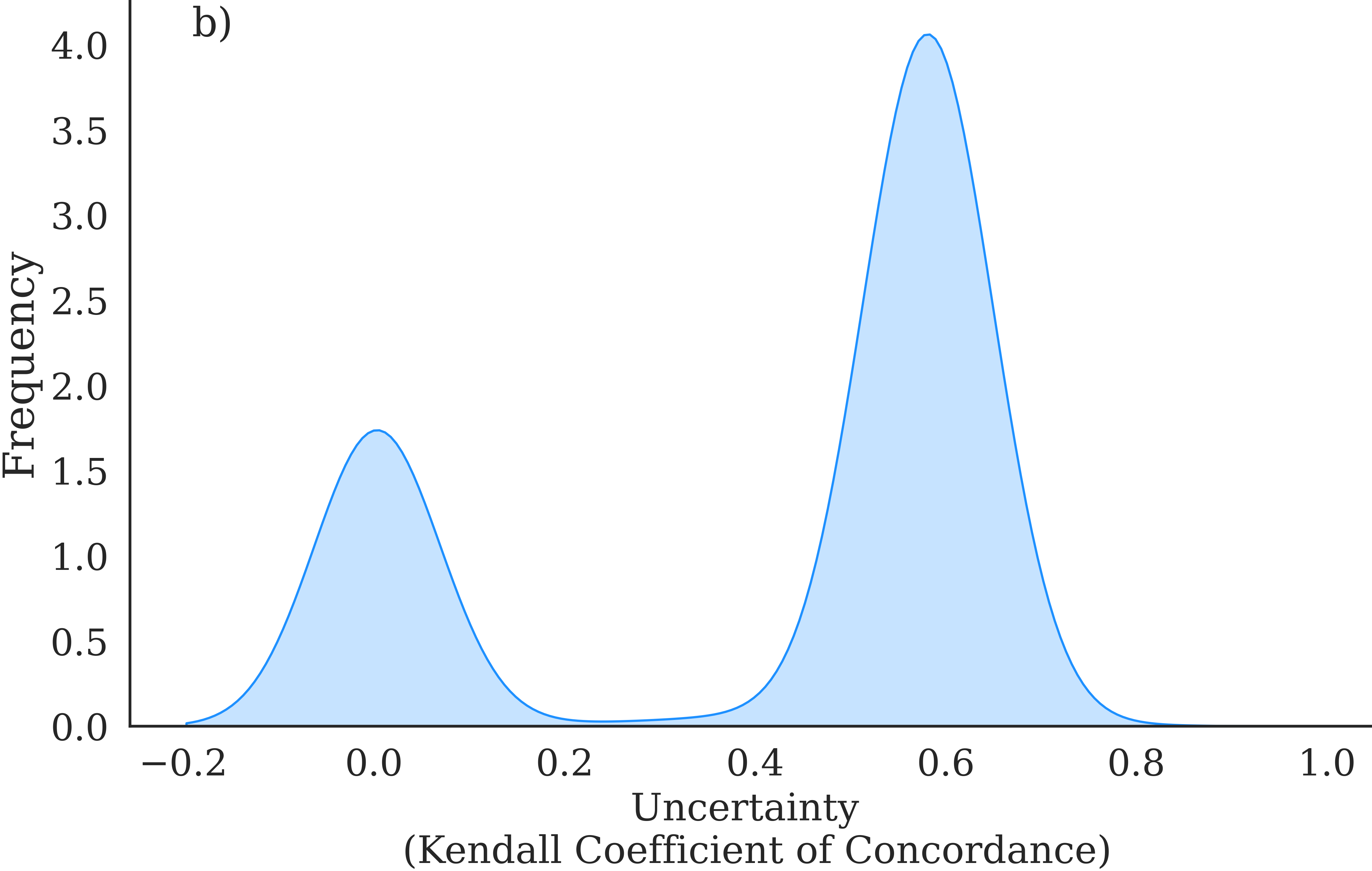} }}%
    \caption{Uncertainty (Kendall Coefficient of concordance) vs Frequency of Instances with corresponding uncertainty for (left) a 3-layer MLP on the PIMA Indian Diabetes dataset, (right) a 16-layer MLP on the synthetic dataset.}%
  \label{fig:diab_unc_freq}
\end{figure*}

\section{Uncertainty Across the Classifier's Feature Space} \label{sec:three}

\begin{table*}[ht]
    \centering
    \caption{Confusion Matrix depicting instances marked "Stable" and "Not Stable" by our algorithm and their corresponding truth labels low (near-zero)/high uncertainty on the Diabetes and Synthetic dataset for LIME and KernelSHAP feature attributions.  }
    \begin{tabular}{c|c||c|c||c|c}
    \hline
     Algorithm & & \multicolumn{2}{|c|}{LIME} & \multicolumn{2}{|c}{KernelSHAP} \\ \hline 
  Dataset  &  & Low Uncertainty   & High Uncertainty & Low Uncertainty &  High Uncertainty  \\ \hline\hline
 
  Diabetes  & Stable & 0.672 & 0.02  & 0.528 & 0.098 \\
    Diabetes  & Not-Stable & 0.019&0.289 & 0.045 &  0.329\\ \hline 

  Synthetic & Stable &     0.264     & 0.022    & 0.205 & 0.01  \\
   Synthetic & Not-Stable &     0.011     & 0.703    & 0.06 &  0.725 \\\hline
    \end{tabular}
    \label{tab:hypothesis1}
\end{table*}

Although there has recently been a lot of interest in quantifying uncertainty of explanation algorithms, studies attempting to do so largely sample a few instances from the classifier's feature space and report aggregrated uncertainty. We, however, think that there might be underlying trends and insights on how the uncertainty of an instance varies across the feature space for a binary classification model. Our aim in this section is to \textit{study how the uncertainty of instances varies across the feature space of a classifier}.

 As a first step towards this, we sample a 1000 instances arbitrarily from the feature space of both the Diabetes and synthetic datasets and compute their uncertainty for LIME explanations using Kendall's coefficient of concordance.  We plot a density graph for both the datasets in Figure \ref{fig:diab_unc_freq}. For both datasets, we observe an interesting phenomenon. We essentially witness two peaks in the Uncertainty density curves for both datasets, one near zero and another at a positive uncertainty value. Instances with feature attribution having near zero uncertainty can be of interest to both earn the trust of end users as well as for use in scientific discovery.
 As a result, we coin the term {\bf stable instances} for such instances. Stable instances can be used as representative instances for global explanations over their more uncertain counterparts. They might also help us mine interesting insights about the dataset, the binary classifier, as well as the explanation algorithm's behaviour while explaining it. As a result it is of interest to diagnose what underlying phenomena makes an instance stable and to come up with an algorithm to generate stable instances.

Here, we focus on diagnosing what makes an instance a stable instance in perturbation-based explanation algorithms. As discussed in the previous section, we find that the nearest decision boundary point ($DBP_x$) has the most influence on the LIME explanation attribution. We find that for most instances, the uncertainty of the instance was significantly less than the uncertainty of its nearest DBP (with $94\%$ accuracy) (Figures \ref{fig:enn_vs_pts_diab_log},\ref{fig:enn_vs_pts_diab_mlp1},\ref{fig:enn_vs_pts_diab_mlp3} in Appendix). We also empirically observe that having multiple DBPs in equal proximity of an instance is a leading cause of high instance uncertainty. As a result, we introduce the term $l_{DBP_k}$ for an instance $x$, which signifies the Euclidean distance between the instance and its $k^{th}$ nearest DBP in the feature space. We hypothesize that for an instance if $l_{DBP_2}$ is larger than $l_{DBP_1}$ by at least $5\%$, the instance is a stable instance. This is motivated by the fact that if the second nearest DBP is at least a threshold distance away from the nearest, it would not influence decision making as much and the explanation would be stable.

To verify our hypothesis, we select 1000 instances sampled uniformly from the feature space of both datasets. We first compute the uncertainty of each instance using Kendall's coefficient of concordance. Following the uncertainty density graphs (Figure \ref{fig:diab_unc_freq}), we term any instance with uncertainty less than $0.02$ in the Diabetes dataset and any instance with uncertainty less than $0.2$ in the synthetic dataset to be stable instances. We then test our hypothesis to compute whether an instance is stable. We sample heavily in the growing spheres algorithm ($n=100,000$) used to compute DBPs. We find that although the growing spheres algorithm is a randomized algorithm, it is deterministic enough to show trends. We report the confusion matrix of our Stable Instance determination algorithm in Table \ref{tab:hypothesis1} for LIME and KernelSHAP on both datasets. Overall, our algorithm has a precision of $89.2\%$ and a recall of $95.1\%$ average over both datasets and algorithms.

The presence of multiple DBPs in close proximity leads to different subsets of them being sampled by perturbation based algorithms at each execution. Since each decision boundary point usually has its own distinct tangent, the incorporation of multiple instances brings in high levels of uncertainty to the LIME algorithm. Following the above, we end up with an algorithm to mine stable instances for a dataset and binary classifier.  This can be incorporated within various global explanation methods that look for representative instances to explain the model's behavior in different regions of the dataset (e.g., SP-LIME \cite{ribeiro2016should}, Guided-LIME~\cite{sangroyaguided}, etc).
We also observe that the fraction of stable instances decreases as the underlying binary classifier's complexity increases.


\begin{figure*}[!t]
    \centering
    \subfloat{{\includegraphics[width=0.45\linewidth]{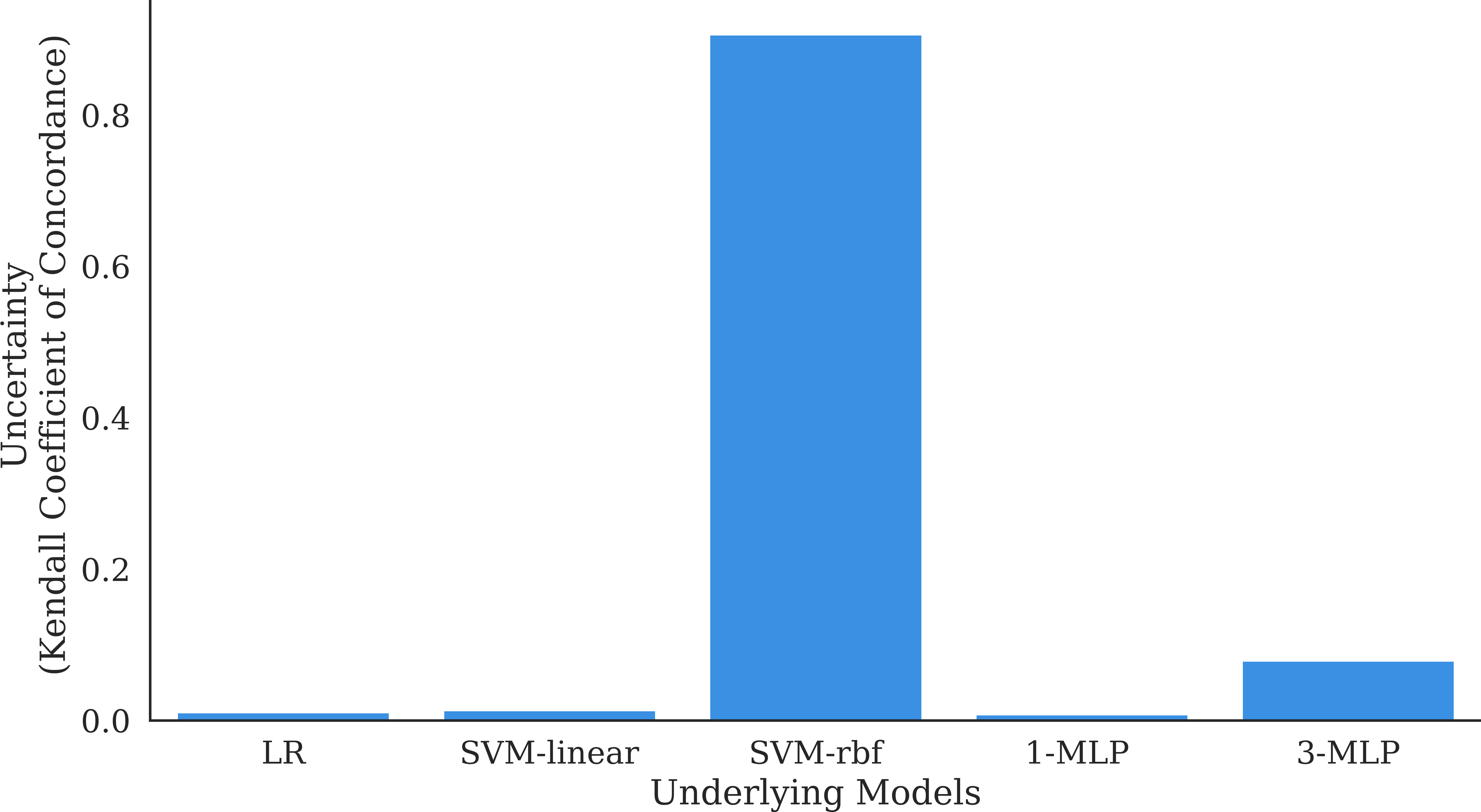} }}%
    \qquad
    \subfloat{{\includegraphics[width=0.45\linewidth]{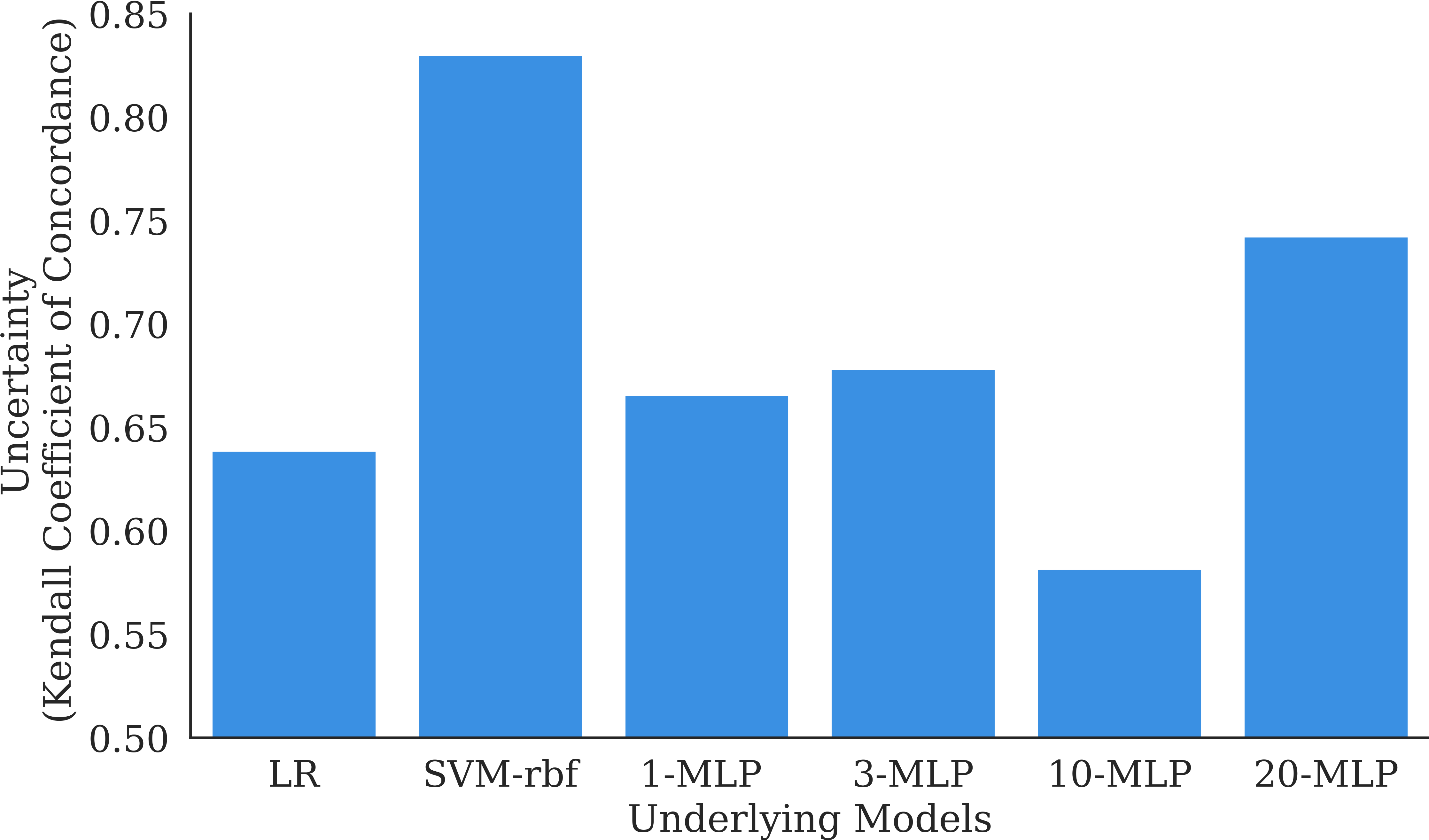} }}%
    \caption{Relative uncertainty amongst various classifiers trained on (a) PIMA diabetes (b) synthetic dataset.}%
  \label{fig:Uncertainty_vs_Model}
\end{figure*}

\begin{algorithm}[!t]
    \caption{ \texttt{ModelComplexityFinder}}
  \small
    \begin{algorithmic}[1]
        \Require Classifier $f$, explainer $g$, dataset $\mathcal D$, number of samples $n$, threshold $m$
        \Procedure{ModelComplexityFinder}{$f, g, \mathcal D, n$}
            \State $p \gets \textsc{RandomSample}(\mathcal{D}, n)$ \Comment Sample $n$ points from $\mathcal D$
            \State $CardinalityArray \gets []$ 
            \ForAll{$x \in p$}
                \State $DBP, l_{DBP} \gets \textsc{NearestDBPs}(f,g,x)$  \Comment Returns DBPs in increasing order of distance to $x$
                \State $c \gets \underset{i}{\text{argmax }} l_{DBP}[i]$ s.t. $l_{DBP}[i] < (1 + m/100) \cdot l_{DBP}[0]$
                \State $CardinalityArray.\textsc{Append}(c)$
            \EndFor
            \State \Return $\frac{1}{n}\Sigma_i CardinalityArray[i]$
        \EndProcedure
    \end{algorithmic}\label{alg:complexityfinder}
\end{algorithm}

\section{Uncertainty vs Underlying model complexity} \label{sec:four}
 Most work in quantifying uncertainty involve experiments on a single underlying classifier. These classifiers are usually available as an API to query and are thus blackbox to the person performing experiments. It is of interest to infer about the complexity of the blackbox classifier, in order to make informed decisions. In this section, our goal is to \textit{study the change in uncertainty of explanation algorithms while computing feature attributions for underlying models of different complexities}.

 We experiment by generating feature attributions for a host of underlying models (Logistic regression, SVM linear kernel, SVM rbf kernel, different MLP setting, etc.) on both the datasets. We use the scikit-learn implementations of each of the above algorithms to learn binary classifiers on the Diabetes and synthetic datasets. We follow by randomly sampling 100 instances from the feature space of each dataset and generating perturbation-based explanations for each of these instances explaining underlying models. We then bootstrap and compute the uncertainty of each of these classifiers using the Kendall's coefficient of concordance. The distribution of uncertainty in feature attributions as a function of the underlying model is visualized in Figure \ref{fig:Uncertainty_vs_Model}. We witness a significant variation (Standard deviation of $0.3524$ and $0.0789$ in PIMA Diabetes and Synthetic datasets respectively) of uncertainty in explanations generated by the same Explanation algorithm (LIME) when explaining different underlying models. We observe that uncertainty increases considerably with the increase in complexity of the underlying model, with small MLP models having least uncertainty and the SVM model with an rbf kernel showing the highest uncertainty. 

In this light, we propose \texttt{ModelComplexityFinder}, an algorithm to quantify the relative ``complexity'' of a blackbox classifier. This ``complexity'' value obtained from \texttt{ModelComplexityFinder} can be incorporated in XAI algorithms to infer more about the blackbox being explained and as a result generate more trustworthy explanations. For example, we can scale the number of perturbations to be sampled in perturbation-based algorithms according to the underlying model's complexity, sampling more densely for complex models to obtain tighter bounds on uncertainty. 

The \texttt{ModelComplexityFinder} algorithm is detailed in Algorithm~\ref{alg:complexityfinder}. We tweak the growing spheres algorithm (Algorithm~\ref{alg:growingSpheres} in the Appendix) to return all nearest decision boundary points within a certain threshold distance from the instance. We then assign a complexity cardinality to each instance based on the number of decision boundary points which lie within $5\%$ of $l_{DBP_1}$. For each classifier, we sample 100 instances uniformly from the feature space and compute the  cardinality of each instance using the tweaked growing spheres algorithm. We average the cardinality of the 100 instances to report the average cardinality complexity of the blackbox model. In this manner,  \texttt{ModelComplexityFinder} allows us to compute the average complexity of a blackbox model before we use standard explanation algorithms to generate explanations for it.

We employ \texttt{ModelComplexityFinder} to estimate the complexity of 5 different underlying models -- Logistic Regression, SVM with a linear Kernel, SVM with an rbf kernel, MLP with 1 hidden layer, and MLP with 3 hidden layers on the Pima Indian Diabetes dataset. All hyper-parameters are kept fixed across the experiments. For the growing spheres algorithm, we set the sampling parameter $n$ to 100,000. We report the results of our \texttt{ModelComplexityFinder} algorithm in Table \ref{tab:complexityfinder}. We observe that according to our algorithm, SVM-rbf has the highest complexity with an average cardinality of $29.8$ while Logistic regression has the lowest with an average cardinality of $3.67$ for the PIMA Diabetes dataset. We evaluate the performance of our proposed algorithm to compute a binary classifier's complexity by calculating the correlation between the ground truth uncertainty (obtained by averaging Kendall's Concordance for those instances) and its calculated complexity. The Pearson correlation coefficient between the model complexity and true model uncertainty is $0.836$, showing that \texttt{ModelComplexityFinder} is a good way to estimate the relative uncertainty of a classifier when all other parameters are fixed.

\begin{table}
    \centering
    \caption{Underlying model's complexity estimated by the \texttt{ModelComplexityFinder} algorithm for various binary classifiers (Logistic Regression (LR), SVM with a linear kernel (SVM-lin) , SVM with an RBF kernel (SVM-rbf), 1 layer MLP, 3 layer MLP) on the Pima India Diabetes and the synthetic dataset. }
    \begin{tabular}{c|c|c|c|c|c}
    \hline 
    Dataset & LR     & SVM-lin & SVM-rbf & 1-MLP & 3-MLP  \\ \hline 
    Diabetes &   3.67 & 4.2 &29.8 & 7.2& 21.6\\ \hline \hline
    Dataset & LR  & SVM-lin & SVM-rbf & 5-MLP & 20-MLP  \\ \hline 
    Synthetic &   1.11 & 1.03 & 1.4 &1.16 & 1.21 \\ \hline
    \end{tabular}
    \label{tab:complexityfinder}
\end{table}


\section{Related Work}
Certain XAI algorithms report uncertainty by default, along with their feature attribution. \citet{schwab2019cxplain} propose CXPlain, which is trained using a causal objective. After training, the model can be used to compute explanations quickly and to estimate uncertainty through bootstrap ensembling. \citet{slack2021reliable}  propose BayesLIME and BayesSHAP, which are Bayesian versions of LIME and KernelSHAP that output credible intervals for the feature importance, capturing the associated uncertainty through confidence intervals. Throughout this study, we treat uncertainty as an emergent behaviour of explainable-AI systems. However, some other works like that of \citet{zhang2019why}, focus on the root cause of uncertainty and  identify three different sources  in LIME-like algorithms, i.e., the randomness of the sampling procedure, variation in the sampling proximity (the kernel function) and variation in explained model credibility across data instances. \citet{schulz2021uncertainty} quantify uncertainty by measuring the ordinal consensus among a set of diverse bootstrapped surrogate explainers. Others like \citet{hill2022explanation} focus on generating an uncertainty estimate combining the decision boundary uncertainty and explanation function uncertainty. A few studies do not focus on uncertainty directly, but throw light on related proxy metrics. \citet{krishna2022disagreement} study the \textit{disagreement problem}, where they work on computing the uncertainty between feature attributions generated by different attribution algorithms for the same classifier and instance.  \citet{alvarezmelis2018robustness} verify the robustness of perturbation-based and saliency-based explanation methods using a local Lipschitz estimate to check how output varies according to a given input. On similar lines, \citet{ghorbani2019interpretation} observe that the interpretations by neural networks are \textit{fragile} and show that two perceptively indistinguishable inputs with the same predicted label can be assigned very different interpretations, via experiments on the MNIST dataset. It is of future interest to study the relation between \textit{uncertainty} and \textit{fragility} metrics for XAI algorithms.

\section{Conclusion and Future Work}
In this work, we addressed a number of important problems under the umbrella of uncertainty in Explainable-AI. We first described our assumptions for uncertainty and enumerated methods to compute uncertainty in XAI. We evaluated the uncertainty of popular explanation methods LIME, Kernel-SHAP, Bayes-LIME, and CXPlain using the described metrics. We next analysed the uncertainty shown by different features within an attribution and proposed an algorithm to achieve negative correlation between a feature attribution and its uncertainty. Next we studied how the uncertainty of explanation vectors varied across the feature space of a model. We highlighted the existence of stable points, diagnosed the reason for their existence and proposed an algorithm to mine a stable instance. Lastly we analysed the impact that the underlying model's complexity had on the uncertainty of explanations and observed that the more complex the underlying model, the higher the resulting explanation uncertainty is. Following this we suggested a proxy metric to estimate an underlying model's complexity in order to set better hyper-parameters in the explanation algorithm. 

Uncertainty of XAI algorithms is a major deterrent in their use in critical real life scenarios. Especially, uncertainty of these explanation algorithms need to be at near zero levels for them to be used to build hypothesis in scientific understanding. Our work summarizes a few issues regarding uncertainty and suggests approaches to solve them for the class of LIME based explanation methods. However the problem of uncertainty in XAI is far from solved. Future work would involve looking into methods to reduce uncertainty in non-perturbation explanation algorithms. The ultimate goal would be to propose explanation algorithms that perform highly with respect to both faithfulness as well as certainty.

\nocite{*}
\bibliography{AAAI24/aaai24}

\appendix

\clearpage

\section{Appendix}

 \begin{figure}[H]
  \centering
  \scalebox{1.2}{
  \includegraphics[width=0.8\linewidth]{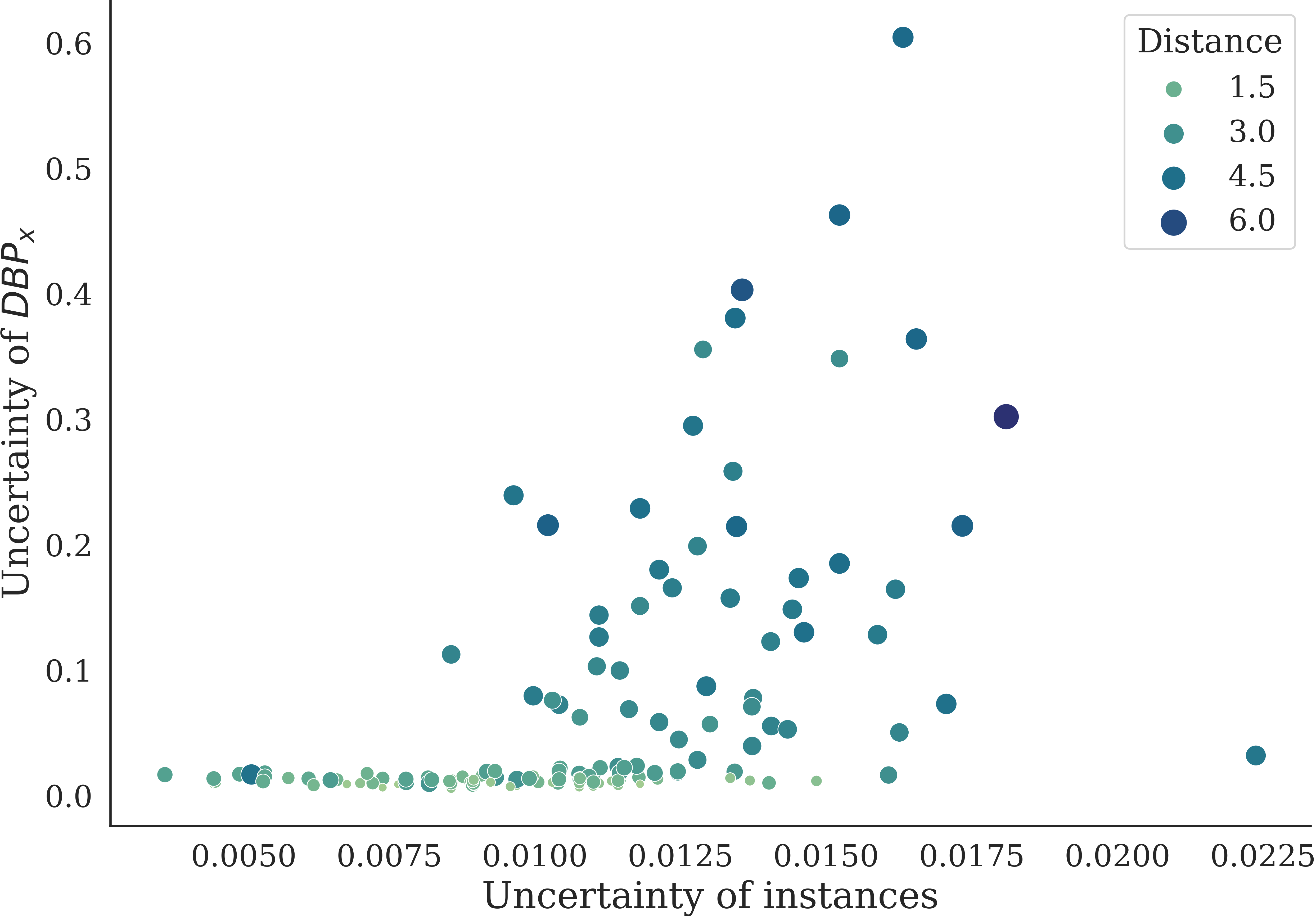}}
  \caption{This figure depicts Uncertainty of instances vs Uncertainty of their closest decision boundary points. The underlying model is Logistic Regression and explanation model is LIME. The points in the plot depicts distance between the instances and their respective DBPs.}
  \label{fig:enn_vs_pts_diab_log}
\end{figure}

\begin{figure}[H]
  \centering
  \scalebox{1.2}{
  \includegraphics[width=0.8\linewidth]{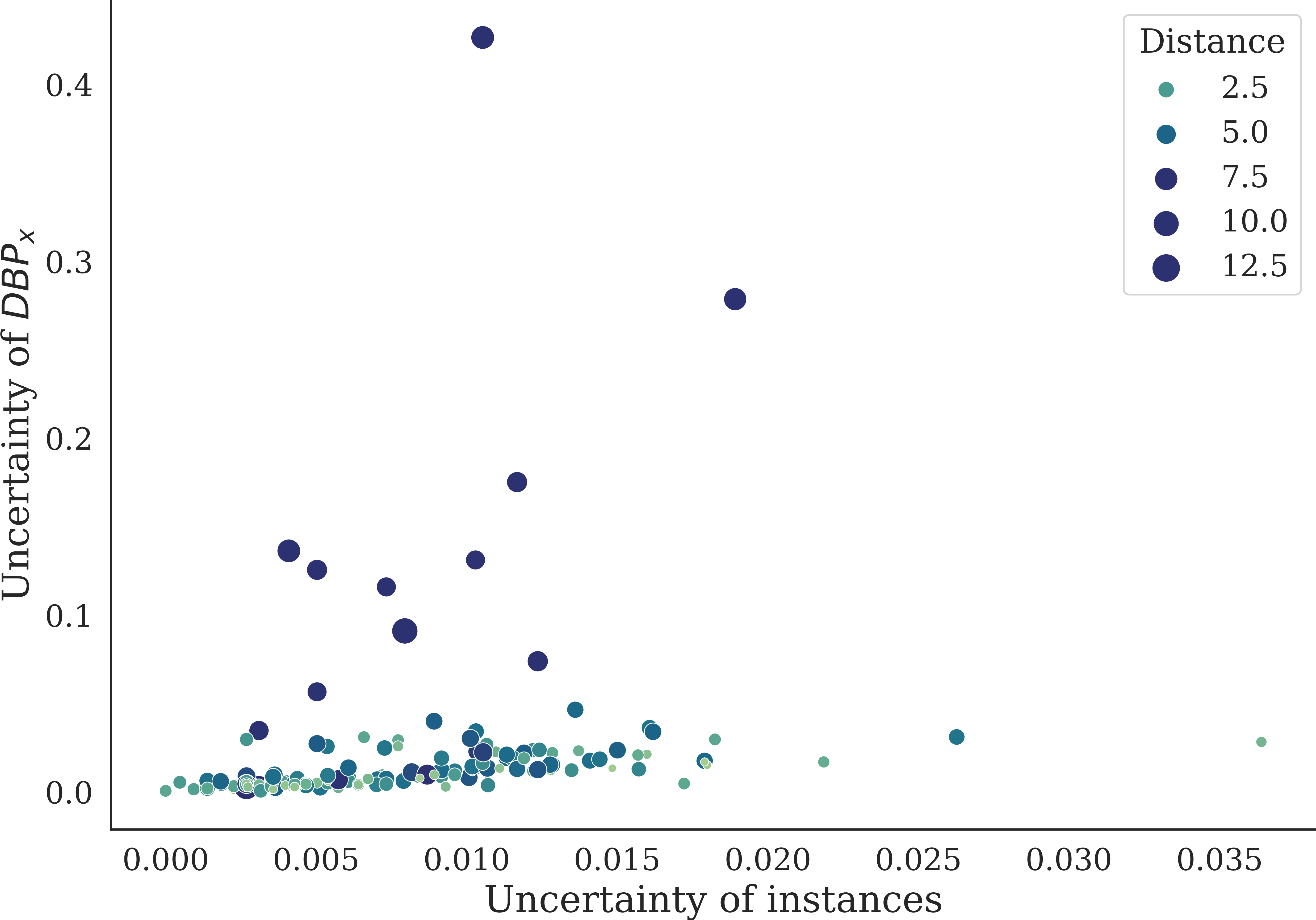}}
  \caption{This figure depicts Uncertainty of instances vs Uncertainty of their closest decision boundary points. The underlying model is MLP with 1 layer and explanation model is LIME. The points in the plot depicts distance between the instances and their respective DBPs.}
  \label{fig:enn_vs_pts_diab_mlp1}
\end{figure}

\begin{figure}[H]
  \centering
  \scalebox{1.2}{
  \includegraphics[width=0.8\linewidth]{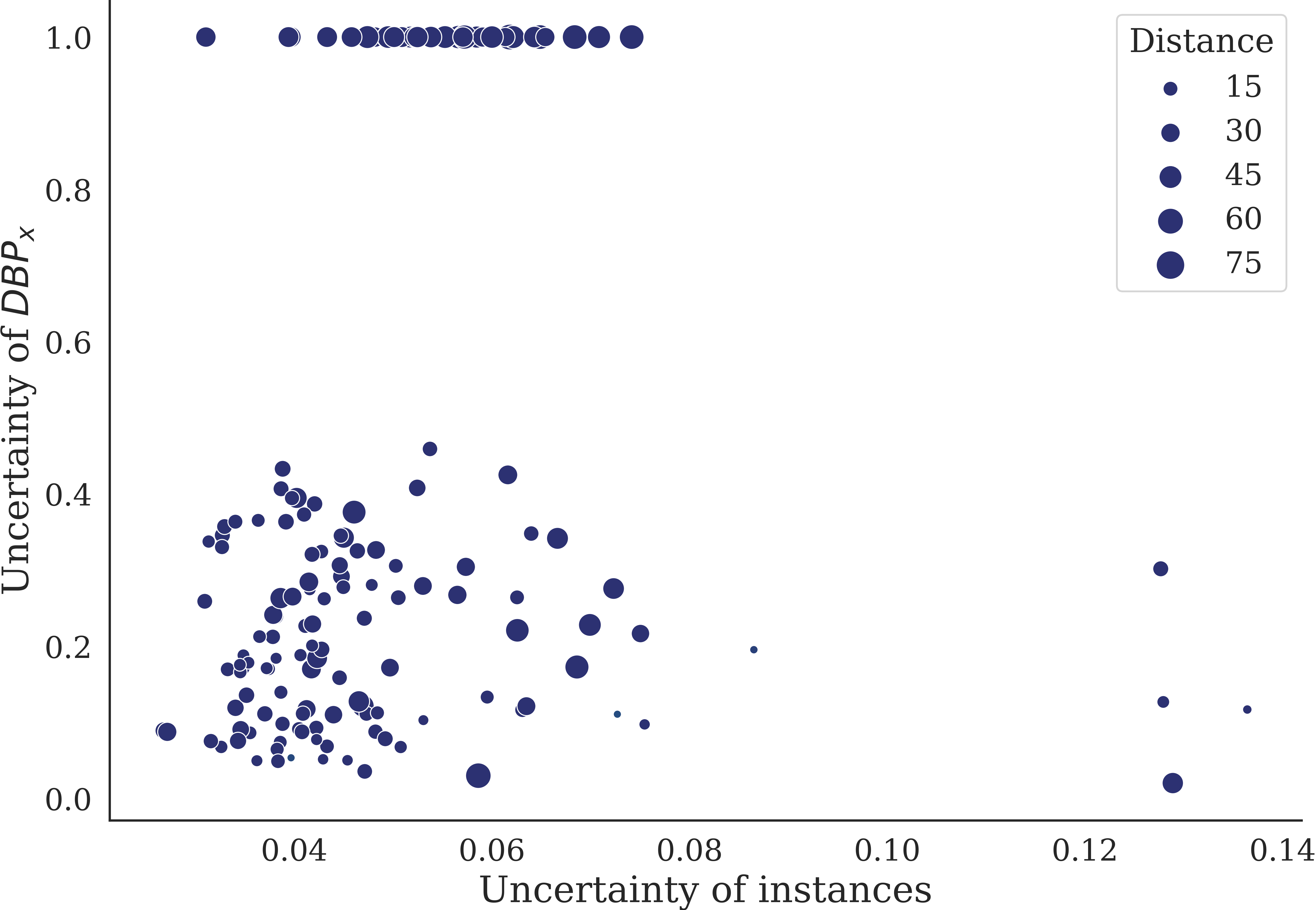}}
  \caption{This figure depicts Uncertainty of instances vs Uncertainty of their closest decision boundary points. The underlying model is MLP with 3 layer and explanation model is LIME. The points in the plot depicts distance between the instances and their respective DBPs.}
  \label{fig:enn_vs_pts_diab_mlp3}
\end{figure}

\begin{figure}[H]
  \centering
  \includegraphics[width=0.8\linewidth]{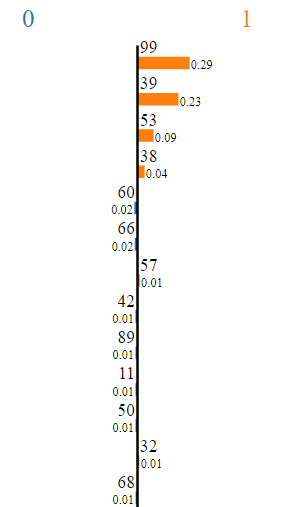}
  \caption{LIME explanation of a prediction made by 16-MLP on an instance of the synthetic dataset.}
  \label{fig:synth_lime}
\end{figure}

Figure \ref{fig:enn_vs_pts_diab_log},\ref{fig:enn_vs_pts_diab_mlp1} and \ref{fig:enn_vs_pts_diab_mlp3} depict the Uncertainty of instances (x-axis) vs Uncertainty of their nearest DBPs (y-axis) for three different binary classification architectures : Logistic Regression, 1 layer MLP and 3 layer MLP respectively. \\

Figure \ref{fig:synth_lime} depicts the LIME explanation for an instance of the synthetic datset. We see that only the top 3 out of 100 features have significant feature attribution and it would be of advantage to make the uncertainty of primary features minimal at the cost of the uncertainty of less important features.

\clearpage

\textbf{Confidence Interval (CI):} This method was used by  \citet{schwab2019cxplain} to compute the confidence interval of an explanation   $CI_{i,\gamma}= [c_{i,\frac{\alpha}{2}}, c_{i,1-\frac{\alpha}{2}} ]$ at confidence level $\gamma = 1- \alpha$ for each assigned feature importance estimate $a_i$.  The width of this confidence interval $u_i = c_{i,1-\frac{\alpha}{2}} - c_{i,\frac{\alpha}{2}}$ is used as a metric to quantify the uncertainty of  $a_i$.
    
\textbf{Confidence Interval + Bootstrapping (CI-bootstrap):} This method was employed by \citet{schwab2019cxplain} to quantify uncertainty of the CXPlain algorithm based on bootstrap resampling. The explanation algorithm is executed $M$ times, and the median of the feature attributions in the ensemble members is assigned as the feature importance of the bootstrap ensemble. This new feature attribution is used to compute confidence intervals using the CI method described above.
    
\textbf{Standard Deviation:} The explanation algorithm is executed $M$ times to compute the standard deviation of each feature's importance values. The values of the features' uncertainty are averaged to compute the uncertainty of an instance's explanation.
    
\textbf{Kendall's Coefficient of Concordance ($W$)}: The method introduced by \citet{kendall1939problem} to quantify the correlation between rankings, was first used by \citet{schulz2021uncertainty} to quantify uncertainty in XAI. Each explanation algorithm's output feature is first assigned a rank based on the relative feature importance in the explanation (1 being the most important). The algorithm is executed $M$ times and a $D$-dimensional cumulative sum of ranks vector (CR) is computed. The final concordance metric $W'$ is computed as $\frac{12\sum_{i=1}^{D} \sum_{d=1}^D CR_d - CR_i}{M^2 (D^3-D)} $. Since this is a concordance metric, $(1-W')$ denotes uncertainty. This is a bounded metric with a range of $[0,1]$.

\textbf{Fleiss' Kappa ($\kappa$):} This is  another measure of rank correlation~\cite{fleiss1973equivalence}. The explanation algorithm is executed $M$ times, and a rank based on feature importance is assigned to each feature in every run. A vector $T$ is computed such that $T_{i,j}$ is the count of the number of times feature $i$  obtains rank $j$. Using this, we compute 
$\overline{P} = \frac{1}{D}\sum_{i=1}^{D} \frac{1}{M^2-M} [ \sum_{j=1}^{D} T_{ij}^2 - M]$ and $\overline{P}_e = \sum_{i=1}^{m} (\frac{\sum_{i=1}^{D}T_{ij}}{\sum_{i=1}^{D} \sum_{j=1}^D T_{ij}})^2$. Then, Kappa $(\kappa)$ score can be calculated as:
$\kappa = \frac{\overline{P} - \overline{P}_e}{1 - \overline{P}_e}$ . This is also a bounded metric with range $[0,1]$. 

We also include comparison with a Random Explanation algorithm as a baseline for control. We random uniformly sample $100$ instances from both of our datasets and generate explanations using the XAI algorithms mentioned above to report the average uncertainty computed by each metric. Explanations are generated to describe the same 3-layer MLP binary classifier for all experiments in this section. Figure \ref{fig:unc100} presents the uncertainty of LIME explanations, averaged over $100$ and $1000$ runs for an instance of the Diabetes dataset. 

For the confidence interval experiments, we set the confidence level to $95\%$. The number of executions ($M$) is set to 100. The number of features ($D$) in the Diabetes dataset is $8$, and that in the synthetic dataset is $100$.

\clearpage

\begin{figure*}[!t]
    \centering
    \subfloat{{\includegraphics[width=0.45\linewidth]{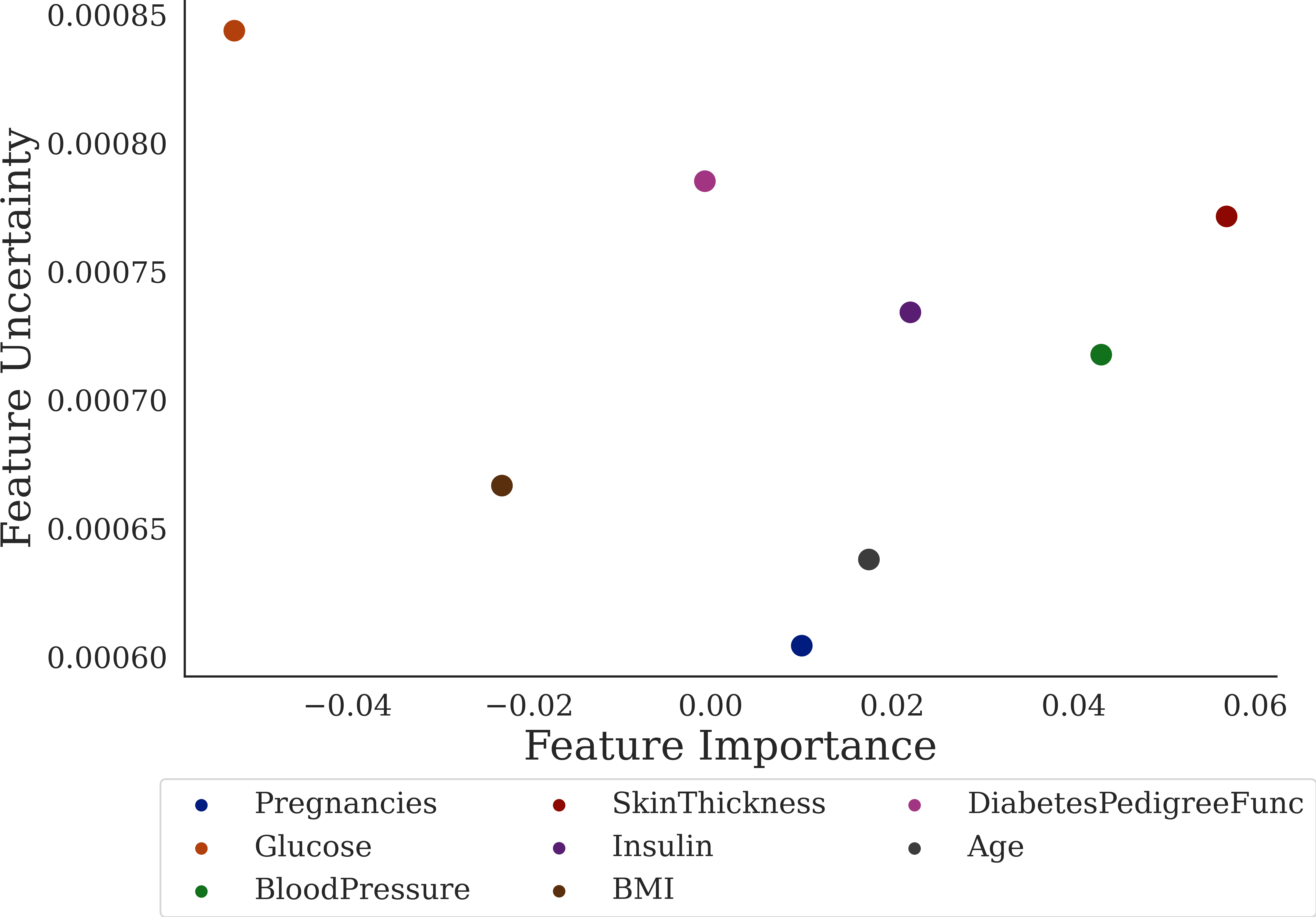} }}%
    \qquad
    \subfloat{{\includegraphics[width=0.45\linewidth]{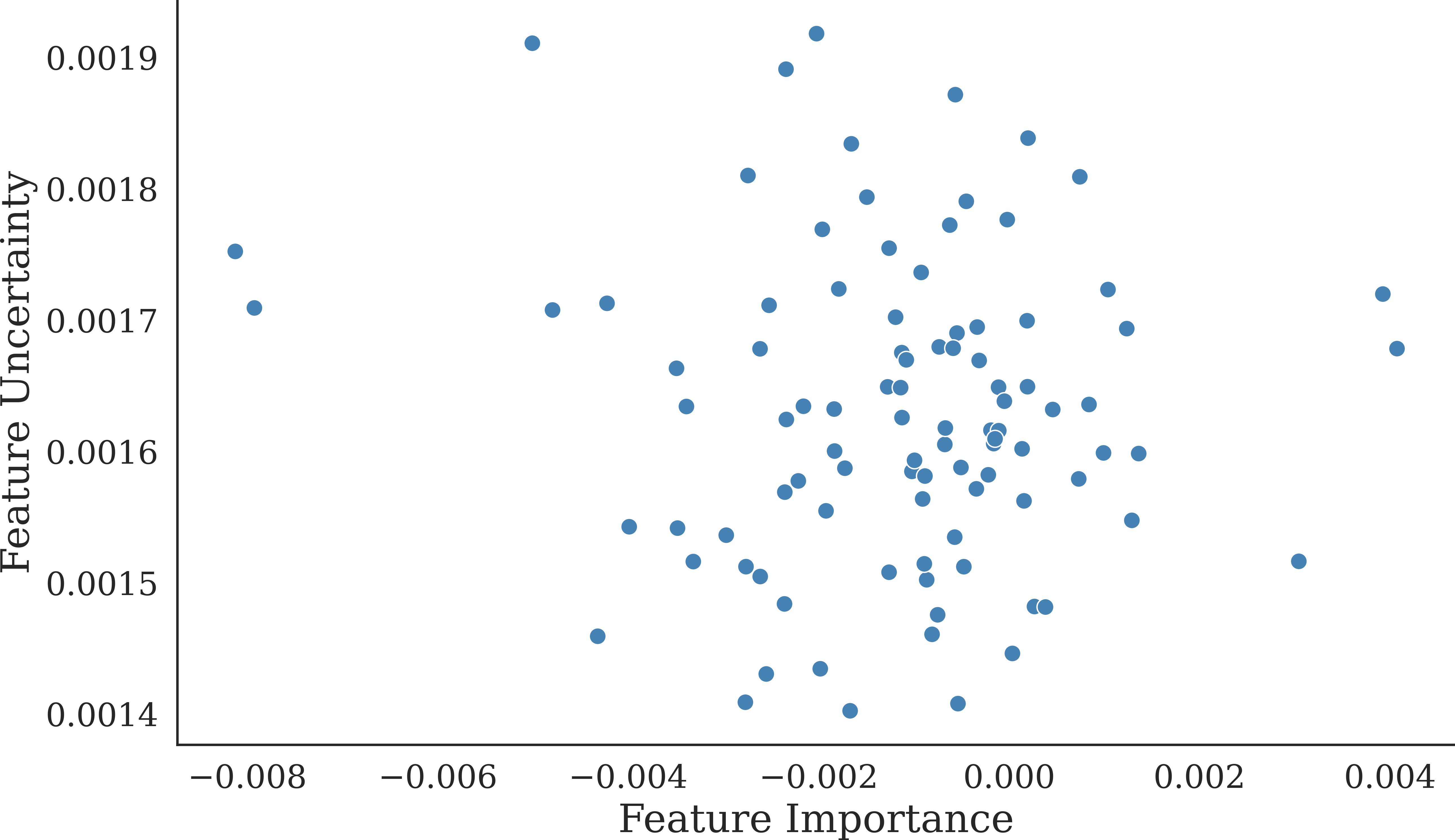} }}%
    \caption{Feature Importance vs Feature Uncertainty for an instance of the (a) Diabetes dataset (b) Synthetic dataset}%
  \label{fig:diab_featimp_featunc}
\end{figure*}

\clearpage

\textbf{Decision Boundary Point ($DBP_x$) :} LIME-based algorithms employ a falling exponential kernel (Equation \ref{eq:3}) based distance metric to compute each perturbation's contribution (Equation \ref{eq:2}) towards the loss function being optimized. 
\begin{align}
    \mathcal{L}(f,g,\pi_x) &= \sum_{z,z' \in Z} \pi_x(z)(f(z)-g(z'))^2 \label{eq:2}\\ 
    \pi_{x}(z)&= \exp(-D(x,z)^2/\sigma^2), \label{eq:3}
\end{align} 
where  $z$ represents the class of perturbed instances, $\pi_x$ describes the locality around $x$, and $\mathcal{L}(f,g,\pi_x)$ is a model of how unfaithful the explanation model $g$ is in neighbourhood of $x$.

This makes finding the nearest Decision Boundary Point (DBP) for an instance of utmost importance as within a set of perturbed instances ($Z$), $DBP_{x}$ is the perturbed instance ($z'$) that will have highest value for $\pi_x(z)(f(z)-g(z'))^2$. Since $\pi_x(z)$ is an exponentially falling kernel, other perturbed instances on the other side of the decision boundary will have little impact on $\mathcal{L}$, as compared to $DBP_{x}$.
In order to compute the decision boundary instance nearest to an instance $x$, Laugel et al.~\cite{laugel2018comparison} propose the Growing Spheres algorithm. They uniformly sample $n$ points in the hyper-dimensional sphere centred at $x$ of radius $\eta$. They initialize $\eta$ with a large value and compute the classification label for each sampled instance as per $f$. The presence of instances with a different classification label than that of $x$ indicates the presence of a decision boundary within the sphere. To find the nearest decision boundary point, they redo the above steps with radius $\eta$ halved and repeat until all sampled points have the same classification label. Next they sample in spherical layers of linearly increasing radii, namely the region between spheres of radius $\eta$ and $2\eta$, then $2\eta$ and $3\eta$, and so on until they encounter a sample instance with classification label difference from $x$. They term this instance $DBP_{x}$. The pseudo-code of this process \cite{laugel2018comparison} is provided in Algorithm~\ref{alg:growingSpheres}. This is an approximate and randomized algorithm due to the high-dimensional sampling involved. We make use of the growing sphere algorithm to compute $DBP_{x}$ in order to analyze the uncertainty displayed by LIME-based algorithms.

We employ concepts from all of the works described above to answer pressing questions and propose interventions regarding the uncertainty of XAI algorithms at the feature level, at the instance level across the feature space as well as at the variation of the underlying model level.

\begin{algorithm}
    \caption{Growing spheres algorithm for computing the Decision Boundary Point nearest to $x$~\cite{laugel2018comparison}}
    \begin{algorithmic}[1]
        \Require $f: \mathbf{x} \rightarrow \{0, 1\}$: a binary classifier , $x \in \mathbf x$: an observation to be explained, $\eta, n$: Hyperparameters
        \Ensure Nearest Decision Boundary Point $e$
        \State Generate $(z_i)_{i \le n}$ uniformly in $\mathrm{SL}(x, 0, \eta)$
        
        \While{$\exists k \in (z_i)_{i \le n}: f(k) \ne f(x) $}
            \State $\eta = \eta / 2$
            \State Generate $(z_i)_{i \le n}$ uniformly in $\mathrm{SL}(x, 0, \eta)$
        \EndWhile
        \State $a_0=\eta, a_1=2\eta$
        \State Generate $(z_i)_{i \le n}$ uniformly in $\mathrm{SL}(x, a_0, a_1)$
        \While{$\Not \exists k \in (z_i)_{i \le n}: f(k) \ne f(x) $}
            \State $a_0 = a_1$
            \State $a_1 = a_1 + \eta$
            \State Generate $(z_i)_{i \le n}$ uniformly in $SL(x, a_0, a_1)$
        \EndWhile
        \State \Return $k$, the $l_2$-closest decision boundary point from $x$
    \end{algorithmic}
    \label{alg:growingSpheres}
\end{algorithm}
\end{document}